\documentclass[10pt,twocolumn,letterpaper]{article}

\usepackage[pagenumbers]{iccv} 

\usepackage[dvipsnames]{xcolor}
\usepackage{multirow}
\usepackage{colortbl}  
\usepackage{xcolor}
\usepackage{array}   
\usepackage{booktabs}
\usepackage{indentfirst}
\usepackage{graphicx}
\usepackage{blindtext}
\usepackage{caption}
\usepackage{subcaption}
\usepackage{ntheorem}
\newtheorem*{thry_algorithm}{}
\usepackage[ruled,vlined]{algorithm2e}
\usepackage{stfloats}
\usepackage[accsupp]{axessibility}

\definecolor{iccvblue}{rgb}{0.21,0.49,0.74}
\usepackage[pagebackref,breaklinks,colorlinks,allcolors=iccvblue]{hyperref}
\definecolor{best}{HTML}{BDE6CD}
\definecolor{second}{HTML}{E4EEBC} 
\definecolor{third}{HTML}{FFF8C5} 

\def\paperID{409} 
\def\confName{ICCV}
\def\confYear{2025}

\title{Stronger, Steadier \& Superior: Geometric Consistency in Depth VFM Forges Domain Generalized Semantic Segmentation}

\author{
Siyu Chen\textsuperscript{\rm 1, 3}\ \ Ting Han\textsuperscript{\rm 2 †}\ \ Changshe Zhang\textsuperscript{\rm 4}\ \ Xin Luo\textsuperscript{\rm 1}\ \ Meiliu Wu\textsuperscript{\rm 3}\ \ Guorong Cai\textsuperscript{\rm 1}\ \ Jinhe Su\textsuperscript{\rm 1 †} 
    \vspace{5pt} \\
    \textsuperscript{\rm 1} Jimei University \textsuperscript{\rm 2} Sun Yat-sen University
    \textsuperscript{\rm 3} University of Glasgow  \textsuperscript{\rm 4} Xidian University \\
}

\begin{document}
\twocolumn[{%
	\renewcommand\twocolumn[1][]{#1}%
	\maketitle%
    \setlength{\abovecaptionskip}{0.1cm}
    \setlength{\belowcaptionskip}{0.1cm}
	\begin{center}
		\centering
        \vspace{-0.6cm}
        \begin{thry_algorithm}\label{fig:figure1}
        \end{thry_algorithm}
    \begin{tabular}{c@{\extracolsep{0.0em}}c@{\extracolsep{0.7em}}c} 
		\includegraphics[width=0.21\textwidth]{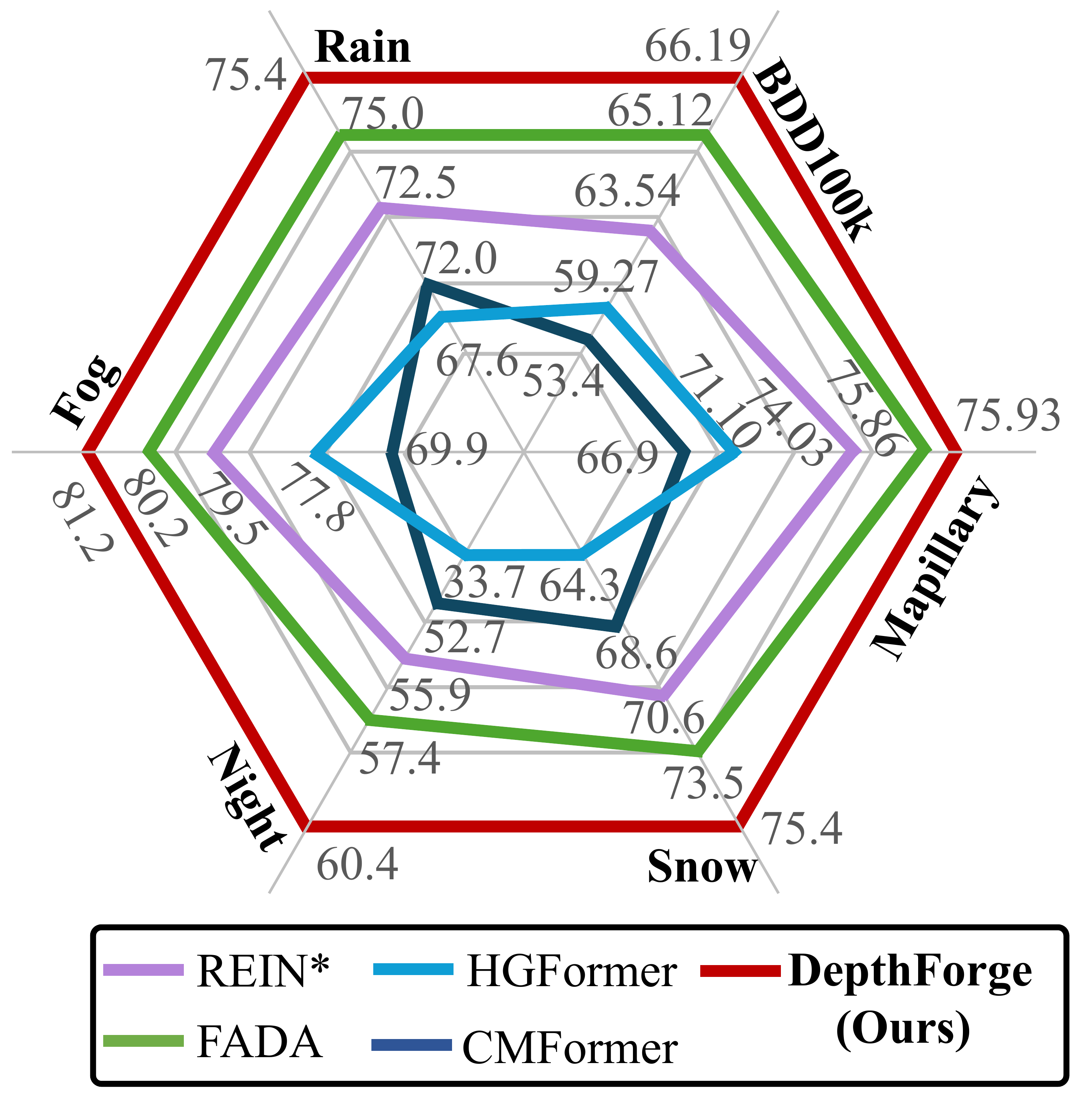}&
		\includegraphics[width=0.37\textwidth]{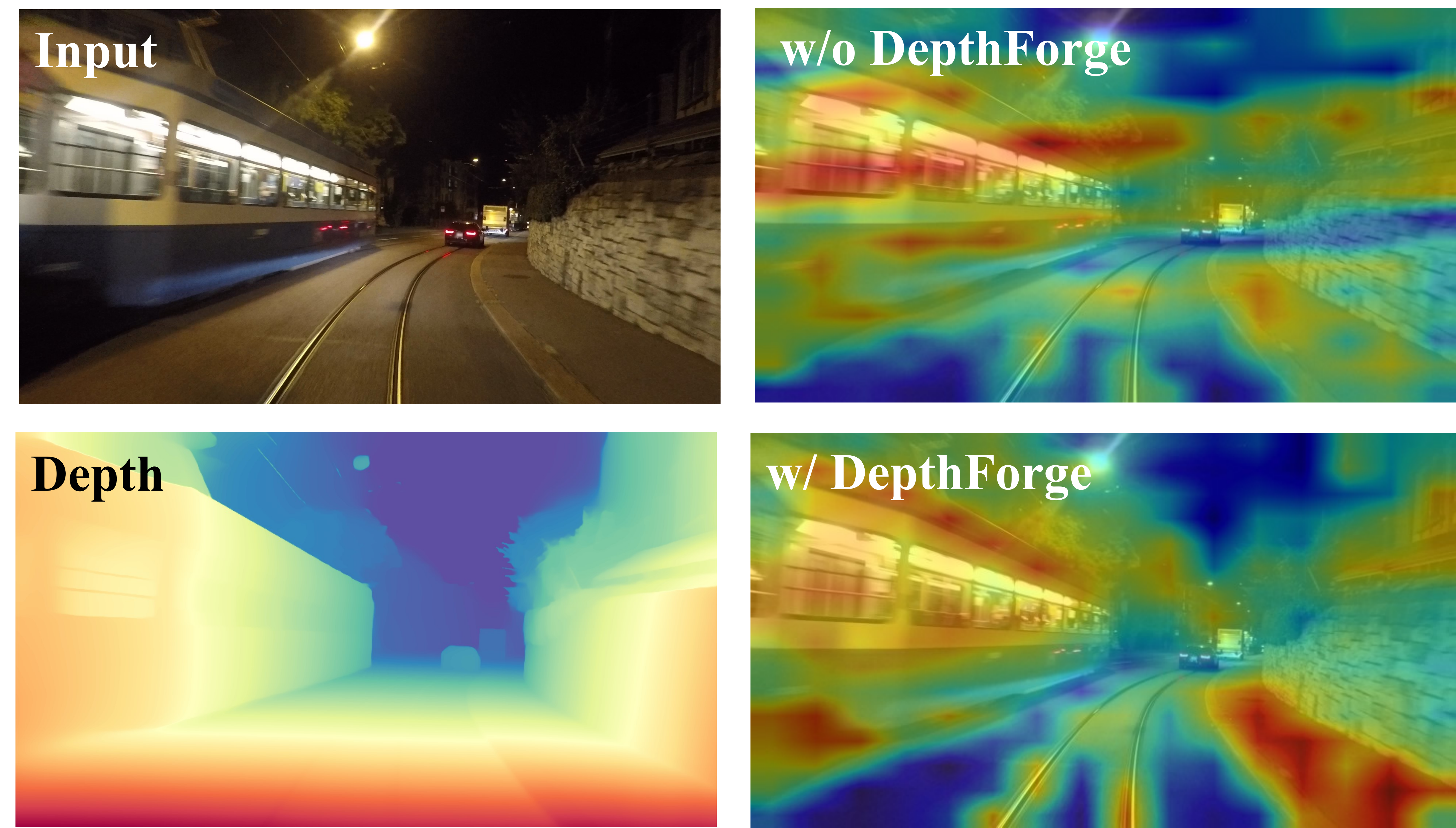}&
		\includegraphics[width=0.37\textwidth]{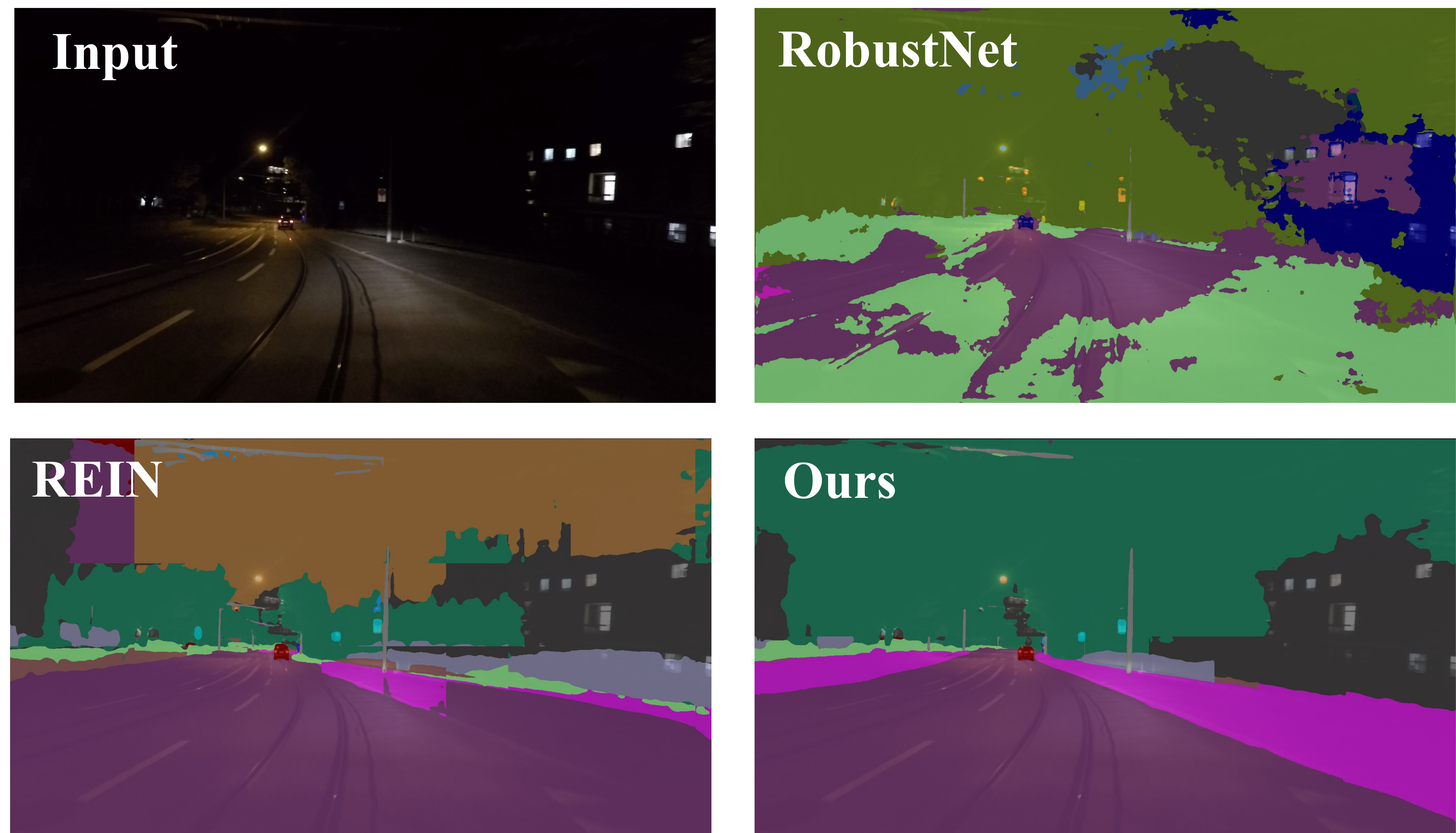}\\
		\footnotesize (a) \textbf{Stronger} Pre-trained Models&	\footnotesize (b) \textbf{Steadier} Visual-Spatial Attention &	\footnotesize (c) \textbf{Superior} Generalization Ability\\
	\end{tabular}
    \captionof{figure}{Existing methods fail to fully exploit the potential of VFMs for DGSS when visual cues are limited or absent. To this end, we introduce a novel and robust fine-tuning strategy \textbf{DepthForge} that leverages depth cues from a depth VFM to optimize visual cues, and employs spatial consistency to enhance feature discriminability. The proposed \textbf{DepthForge} achieves (a) \textbf{stronger} performance under extreme conditions, owing to (b) \textbf{steadier} visual-spatial attention, and thus delivers (c) \textbf{superior} generalization ability.} \label{fig: motivation}
    \end{center}     
}]
\begin{abstract}
Vision Foundation Models (VFMs) have delivered remarkable performance in Domain Generalized Semantic Segmentation (DGSS). However, recent methods often overlook the fact that visual cues are susceptible, whereas the underlying geometry remains stable, rendering depth information more robust. In this paper, we investigate \textbf{the potential of integrating depth information with features from VFMs}, to improve the geometric consistency within an image and boost the generalization performance of VFMs. We propose a novel fine-tuning DGSS framework, named \textbf{DepthForge}, which integrates the visual cues from frozen DINOv2 or EVA02 and depth cues from frozen Depth Anything V2. In each layer of the VFMs, we incorporate \textbf{depth-aware learnable tokens} to continuously decouple domain-invariant visual and spatial information, thereby enhancing depth awareness and attention of the VFMs. Finally, we develop a depth refinement decoder and integrate it into the model architecture to adaptively refine multi-layer VFM features and depth-aware learnable tokens. Extensive experiments are conducted based on various DGSS settings and five different datasets as unseen target domains. The qualitative and quantitative results demonstrate that our method significantly outperforms alternative approaches with \textbf{stronger performance, steadier visual-spatial attention, and superior generalization ability}. In particular, DepthForge exhibits outstanding performance under extreme conditions (e.g., night and snow). Code is available at \href{https://github.com/SY-Ch/DepthForge}{https://github.com/SY-Ch/DepthForge}.
\end{abstract}    
\vspace{-5pt}
\section{Introduction}\label{sec:intro}

Domain Generalized Semantic Segmentation (DGSS) aims to improve prediction accuracy across multiple unseen domains without requiring access to their data, thereby ensuring robust generalization for practical applications \cite{Yue_2021_CVPR,9879987,10018569}. One common approach involves decomposing the learned features into domain-invariant and domain-specific components \cite{Tang_Gao_Zhu_Zhang_Li_Metaxas_2021,Xu_Yao_Jiang_Jiang_Chu_Han_Zhang_Wang_Tai_2022}. This method aims to achieve robustness against domain variations by isolating content-related factors from those that vary between domains. A second strategy employs meta-learning techniques \cite{Kim_Lee_Park_Min_Sohn_2022}, which focus on training models to generalize effectively across domains by learning more adaptable representations, often leveraging shared structures across different datasets.

Recently, VFMs have emerged as a promising approach, primarily owing to their strong generalization capabilities, which are enabled by pre-trained on large-scale datasets \cite{CLIP,EVA02,SAM}. Fine-tuning these models has become increasingly popular due to their cost-effectiveness and superior performance \cite{FADA,REIN}. However, a significant challenge remains: \textbf{\textit{what constitutes robust feature information for domain generalization, and how can such robust information be effectively extracted?}} Previous methods predominantly utilize VFMs based on RGB images. However, the inherent limitations of RGB prevent those methods from adapting to variable conditions, especially in scenarios such as nighttime, overexposure, snow, or fog, where visual cues are not clearly present. This clearly fails to meet the requirements of domain-generalized semantic segmentation.

In this paper, we introduce \textbf{DepthForge}, a robust fine-tune and optimization strategy for domain generalized semantic segmentation, as shown in Fig.~\ref{fig: motivation}. Specifically, we employ two frozen VFMs: Depth Anything V2\cite{depth_anything_v2,depth_anything_v1} is dedicated to extract depth information, and DINOv2\cite{DINOV2} or EVA02\cite{EVA02} are used to process RGB visual features. We enhance the generalization of the visual features at each layer of the VFMs by integrating the corresponding depth cues. More importantly, we introduce \textbf{depth-aware learnable tokens} that enables the model to learn spatial structure invariance, thereby ensuring consistent performance across different domains. The depth and visual features extracted from frozen VFMs are used to optimize the representation of our learnable tokens, thus producing more robust visual-spatial attention features than those based solely on visual cues and forging geometric consistency, as shown in Fig.~\ref{fig: motivation}(b). However, we observe the frozen VFMs only provide static features, they are incapable of refining the feature maps during training. This limitation leads to persistent errors in the learnable tokens, causing the optimization process to drift in incorrect directions that are difficult to rectify. To this end, we propose a \textbf{depth refinement decoder} that dynamically adjusts the multi-scale features, combining prior and dynamic features to establish high-quality paired feature relationships at different depth spaces.

Our core contribution is to demonstrate that robust depth features and learnable tokens can be effectively applied to improve model performance in Domain Generalized Semantic Segmentation (DGSS). To the best of our knowledge, the proposed DepthForge is the first depth-aware framework specifically designed for DGSS. The effectiveness of our design is verified by experiments on widely used Cityscapes, ACDC, Mapillary, BDD100k, and GTA5 datasets. In addition, the individual components of our design are also verified by extensive experiments. Extensive experiments have demonstrated that our method significantly improves DGSS performance, as shown in Fig.~\ref{fig: motivation}(c). Specifically, in the \textit{GTA$\rightarrow$  Cityscapes + BDD100k + Mapillary} setting, we observed approximately \textcolor{red}{\textbf{+4\%}} significant improvement in mIoU, particularly the improvement in extreme scenes is close to \textcolor{red}{\textbf{+5\%}}. Our main contributions are summarized as follows:
\begin{itemize}
    \item We propose a stronger DepthForge learning framework to fine-tune VFMs combined with visual and depth cues for domain-generalized semantic segmentation.
    \item A depth-awareness learnable tokens are introduced to focus on spatial relationship that acquires steadier and more robust visual-spatial attention. This strategy enables DepthForge to enhance the discriminability and refine features at an instance level within each layer. 
    \item Depth Forge highlights the impressive generalization ability, surpassing existing DGSS methods by a large margin, especially in extreme scenes with limited and absent visual cues.
\end{itemize}

\begin{figure}[!t]
    \centering
    \setlength{\abovecaptionskip}{0.1cm}
    \setlength{\belowcaptionskip}{-0.3cm}
    \includegraphics[width=1\linewidth]{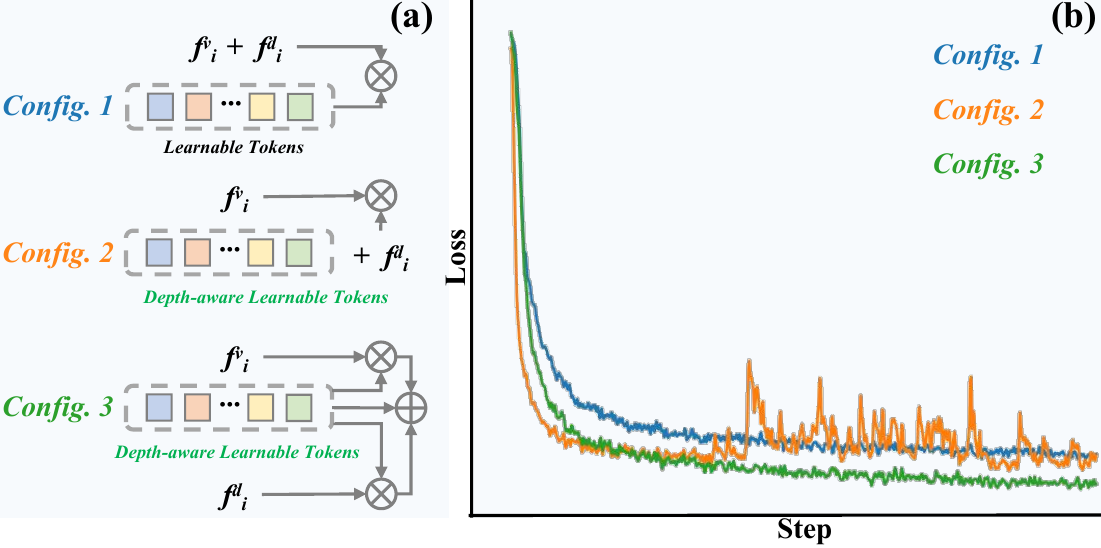}
    \caption{Different configurations of incorporating depth cues into visual information, where \textcolor[HTML]{1C86EE}{Config.1: adding $f^{d}_{i}$ to $f^{v}_{i}$}, or \textcolor[HTML]{FF6347}{Config.2: incorporating $f^{d}_{i}$ into the learnable token $T_{i}$}, and \textcolor[HTML]{2E8B57}{Config.3: our DepthForge}, respectively.}
    \label{fig: loss}
\end{figure}

\section{Related Work}
\label{sec:related}
\subsection{Domain Generalized Semantic Segmentation}

Domain Generalized Semantic Segmentation (DGSS) aims to enhance the generalization capability of segmentation models to unseen domains. Recent methods primarily focus on reducing feature distribution discrepancies across domains and learning more domain-invariant representations. Adversarial training \cite{Yue_2021_CVPR}, meta-learning \cite{NEURIPS2021_b0f2ad44,9577934}, and self-supervised learning \cite{9879121,9879934} have been extensively explored to improve generalization performance. However, these methods often rely on constructing cross-domain contrastive constraints or employing sophisticated training strategies, limiting their generalization improvements and complicating the training process. The emergence of Vision Foundation Models (VFMs), trained via large-scale image-text \cite{CLIP} or image-image contrastive learning \cite{DINOV2}, has opened a promising avenue for DGSS. VFMs capture universal and rich visual representations beneficial for reducing domain biases. Recent approach \cite{REIN,FADA} leverages pre-trained VFMs by employing their frozen features as powerful generalized backbones to extract and refine learnable tokens, leading to significant improvements in model performance. Nevertheless, current DGSS approaches predominantly utilize RGB-only data, leaving the models susceptible to variations in lighting, texture, and spatial arrangements across different domains, thus constraining their full potential for generalization.

\begin{figure*}[!t]
    \centering
    \setlength{\abovecaptionskip}{0.1cm}
    \setlength{\belowcaptionskip}{-0.2cm}
    \includegraphics[width=1\linewidth]{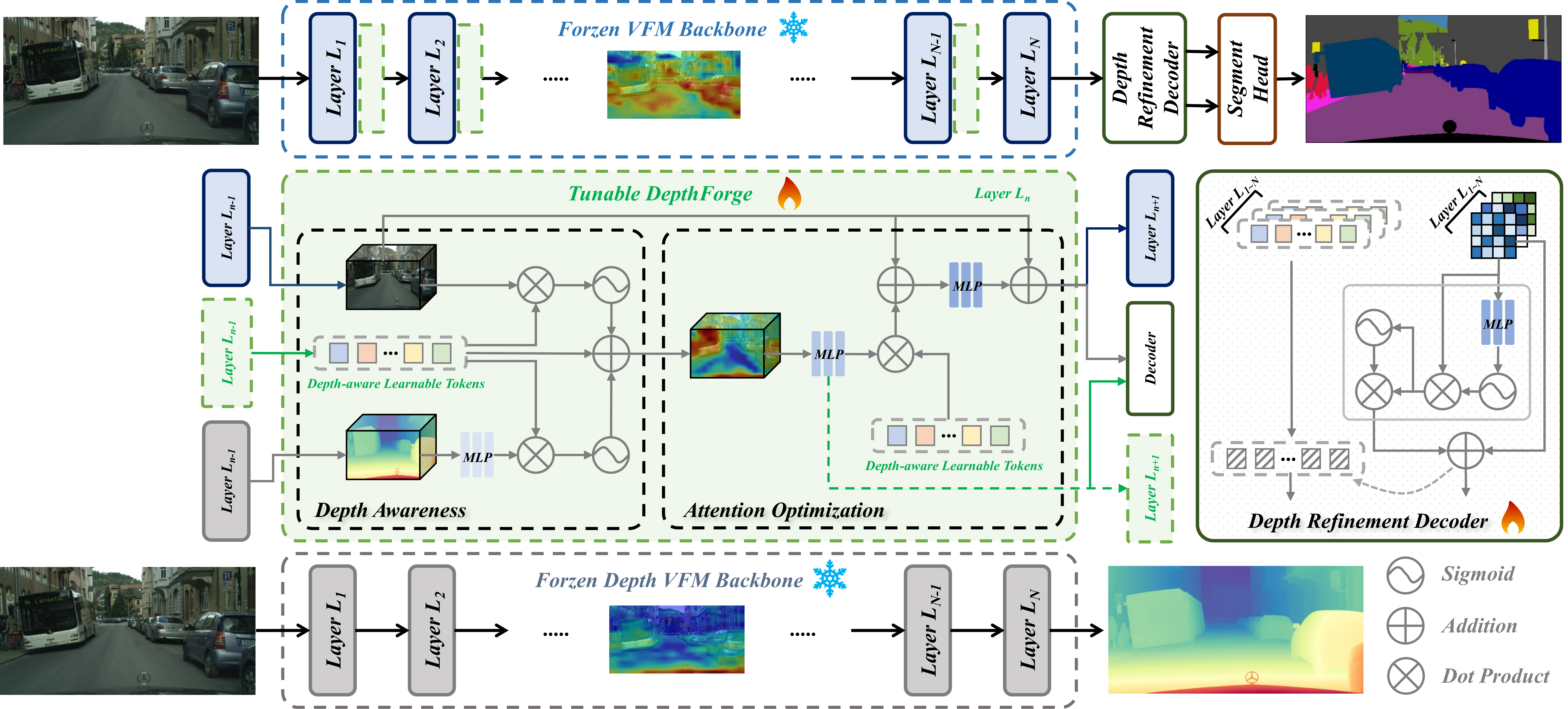}
    \caption{DepthForge framework. It incorporates the features from frozen \textcolor[HTML]{2E75B6}{VFM} and \textcolor{gray}{depth VFM} into our \textcolor[HTML]{00B050}{\textbf{Depth-aware Learnable Tokens}} to generate enhanced features $f_{i+1} = f^{v}_{i} + DepthForge(f^{v}_{i}, f^{d}_{i}) $ for each layer. In our tunable DepthForge, We first design a Depth Awareness module that integrates visual and depth cues into learnable tokens. Next, we propose an Attention Optimization module to strengthen the discriminability of instances across different spatial locations. Finally, multi-scale features are fused in Depth Refinement Decoder to adapt to various scales and spatial connections. \textbf{Our DepthForge establishes direct connections to different instances in real-world locations, facilitating both feature refinement and improved discriminability.}  }
    \label{fig: framework}
\end{figure*}

\subsection{RGB-D Segmentation}

Combining RGB with depth information has shown substantial benefits across various visual tasks \cite{han2024epurate,chen2025hspformer,chen2024trans,rizzoli2024source, chen2024depth}. In object detection, RGB-D effectively mitigates limitations inherent to RGB data under complex scenes by providing robust geometric and spatial cues, thereby enhancing localization accuracy and model robustness \cite{9493207,9424966}. In semantic segmentation, depth information assists models in capturing spatial structures and geometric contours of objects, alleviating issues related to imprecise object boundary delineation. Particularly in scenarios with blurred boundaries, insufficient illumination, or texture scarcity, the incorporation of depth information substantially improves segmentation accuracy and stability \cite{10185116,10252155,10231003,Du_2024_CVPR}. Recent studies have further indicated that even in domain generalization tasks, integrating RGB and depth features consistently improves the model's generalization performance on unseen domains \cite{10288539,Rizzoli_2024_WACV,10268352,10160742}. Therefore, exploring the integration of depth information to address the limitations of RGB-only approaches represents a promising and essential research direction within DGSS.

\section{Preliminary}

To achieve better generalization performance, existing approaches opt to fine-tune VFMs in a parameter-efficient manner. One mainstream idea is to refine and propagate features from frozen VFM layers to subsequent ones. Specifically, given a pre-trained VFM composed of $N$ sequential layers $\{L_{1}, L_{2}, ..., L_{N} \}$, where each layer is associated with a weight matrix $W_{i} \in \mathbb{R}^{c \times c}$ and the frozen output features of each layer are denoted as $f_{i}$ for $i=1, 2, ..., N$. If we have a learnable weight matrix as $ \Delta W_{i} \in \mathbb{R}^{c \times c}$, the feature propagation from layer $L_{i}$ to $L_{i+1}$ is then formulated as:
\begin{equation}
    f_{i+1} = W_{i}f_{i} + \Delta W_{i}f_{i}.
\end{equation}

Recent REIN transfers the learnable matrix $\Delta W_{i}$ into a learnable token $T_{i}$:
\begin{equation}
    f_{i+1} = W_{i}f_{i} + \epsilon(T_{i}(W_{i}f_{i})),
\end{equation}
where $\epsilon(\cdot)$ denotes the MLP function, and $T_{i} \in \mathbb{R}^{m \times c}$ with a tokn length $m$. REIN enables each token to connect better with instances in an image. However, we observe that since the updated and frozen features are limited to remain similar, this correlation is fragile and the capacity for feature refinement is limited.

This prompted us to explore \textbf{what kind of information could improve discrimination in instances and refine the feature representations.}  Drawing inspiration from multimodal learning, we observe that depth features provide a spatial relationship bias for corresponding pixels in visual features, thereby compensating for the loss of crucial visual cues in DGSS. We assume that incorporating a pre-trained depth VFM into the pre-trained VFMs will alleviate attention bias and significantly enhance scene understanding, particularly under extreme conditions.

We use the frozen features $f^{v}_{i}$ from VFM for visual embedding, and the frozen features $f^{d}_{i}$ from depth VFM for depth embedding. A straightforward thought might simply \textcolor[HTML]{1C86EE}{add $f^{d}_{i}$ to $f^{v}_{i}$}, or \textcolor[HTML]{FF6347}{incorporate $f^{d}_{i}$ into the leanable token $T_{i}$}. However, these methods lack flexibility and introduce significant inter-layer feature differences that prevent convergence, as shown in Fig.~\ref{fig: loss}. In contrast, we propose \textcolor[HTML]{2E8B57}{\textbf{DepthForge}} to adaptively fuse visual and depth cues, denoted as:
\begin{equation}
    f_{i+1} = W^{v}_{i}f_{i} + \varepsilon(\epsilon^{v}(T_{i}(W^{v}_{i}f^{v}_{i})) + \epsilon^{d}(T_{i}(W^{d}_{i}f^{d}_{i}))),
\end{equation}
where $W^{v}_{i}$ and $W^{d}_{i}$ represent the weight matrix from VFMs and depth VFM, respectively. $\varepsilon$ denotes the adaptation of the visual and depth signals. This approach allows us to more effectively utilize the powerful capacities of multi-modal VFMs to forge robust refined features and instance discriminability.

\begin{table}[!t]
    \centering
    \scriptsize
    \setlength{\abovecaptionskip}{0.1cm}
    \setlength{\belowcaptionskip}{-0.2cm}
    \setlength{\tabcolsep}{2pt}
    \renewcommand{\arraystretch}{1.1}
    \resizebox{0.43\textwidth}{!}{
    \begin{tabular}{c|c|cccc}
        \toprule
        \textbf{Method} & \textbf{Proc. \& Year} & \textbf{Snow} & \textbf{Night} & \textbf{Fog} & \textbf{Rain} \\
        \midrule
        \multicolumn{2}{l}{\textit{ResNet based:}} \\
        IBN~\cite{IBN} & ECCV2018 & 49.6 & 21.2 & 63.8 & 50.4 \\
        Itenorm~\cite{Itenorm} & CVPR2019 & 49.9 & 23.8 & 63.3 & 50.1 \\
        IW~\cite{IW} & CVPR2019 & 47.6 & 21.8 & 62.4 & 52.4 \\
        RobustNet~\cite{ISW} & CVPR2021 & 49.8 & 24.3 & 64.3 & 56.0 \\
        WildNet~\cite{WildNet} & CVPR2022 & 28.4 & 12.7 & 41.2 & 34.2 \\
        \midrule
        \multicolumn{2}{l}{\textit{Transformer based:}} \\
        HGFormer~\cite{HGFormer} & CVPR2023 & 68.6 & 52.7 & 69.9 & 72.0 \\
        CMFormer~\cite{CMFormer} & AAAI2024 & 64.3 & 33.7 & 77.8 & 67.6 \\
        \midrule
        \multicolumn{2}{l}{\textit{VFM based:}} \\
        REIN~\cite{REIN} & CVPR2024 & 70.6\cellcolor[HTML]{FFF8C5} & 55.9\cellcolor[HTML]{FFF8C5} & 79.5\cellcolor[HTML]{FFF8C5} & 72.5\cellcolor[HTML]{FFF8C5} \\
        FADA~\cite{FADA}  & NeurIPS2024 & 73.5\cellcolor[HTML]{E4EEBC} & 57.4\cellcolor[HTML]{E4EEBC} & 80.2\cellcolor[HTML]{E4EEBC} & 75.0\cellcolor[HTML]{E4EEBC} \\
        \textbf{DepthForge(Ours)} & - & \textbf{75.4}\cellcolor[HTML]{BDE6CD} & \textbf{60.4}\cellcolor[HTML]{BDE6CD} & \textbf{81.2}\cellcolor[HTML]{BDE6CD} & \textbf{75.4}\cellcolor[HTML]{BDE6CD} \\
        \bottomrule
    \end{tabular}
    }
    \caption{Performance comparison between our DepthForge and existing DGSS methods under \textit{Citys. $\rightarrow$ ACDC with Snow, Night, Fog, Rain} generalization setting. Top three results are highlighted as \colorbox[HTML]{BDE6CD}{\textbf{best}}, \colorbox[HTML]{E4EEBC}{second}, and \colorbox[HTML]{FFF8C5}{third}, respectively. (\%)}
    \label{tab: C2A}
\end{table}

\begin{table}[!t]
    \centering
    \scriptsize
    \setlength{\abovecaptionskip}{0.1cm}
    \setlength{\belowcaptionskip}{-0.1cm}
    \setlength{\tabcolsep}{4pt}
    \renewcommand{\arraystretch}{1.1}
    \resizebox{0.48\textwidth}{!}{
    \begin{tabular}{c|c|ccc}
        \toprule
        \textbf{Method} & \textbf{Proc. \& Year} & \textbf{BDD100k} & \textbf{Mapillary} & \textbf{GTA} \\
        \midrule
        \multicolumn{2}{l}{\textit{ResNet based:}} \\
        IBN~\cite{IBN} & ECCV2018 & 48.56 & 57.04 & 45.06 \\
        Itenorm~\cite{Itenorm} & CVPR2019 & 49.23 & 56.26 & 45.73 \\
        IW~\cite{IW} & CVPR2019 & 48.49 & 55.82 & 44.87 \\
        RobustNet~\cite{ISW} & CVPR2021 & 50.73 & 58.64 & 45.00 \\
        DRPC~\cite{DRPC} & ICCV2019 & 49.86 & 56.34 & 45.62 \\
        GTR~\cite{gtrltr} & TIP2021 & 50.75 & 57.16 & 45.79 \\
        SHADE~\cite{SHADE} & ECCV2022 & 50.95 & 60.07 & 48.61 \\
        SAW~\cite{SAW} & CVPR2022 & 52.95 & 59.81 & 47.28 \\
        WildNet~\cite{WildNet} & CVPR2022 & 50.94 & 58.79 & 47.01 \\
        BlindNet~\cite{BlindNet} & CVPR2024 & 51.84 & 60.18 & 47.97 \\
        \midrule
        \multicolumn{2}{l}{\textit{Transformer based:}} \\
        HGFormer~\cite{HGFormer} & CVPR2023 & 53.40 & 66.90 & 51.30 \\
        CMFormer~\cite{CMFormer} & AAAI2024 & 59.27 & 71.10 & 58.11 \\
        \midrule
        \multicolumn{2}{l}{\textit{VFM based:}} \\
        REIN~\cite{REIN} & CVPR2024 & 63.54\cellcolor[HTML]{FFF8C5} & 74.03\cellcolor[HTML]{FFF8C5} & 62.41\cellcolor[HTML]{FFF8C5} \\
        FADA~\cite{FADA}  & NeurIPS2024 & 65.12\cellcolor[HTML]{E4EEBC} & 75.86\cellcolor[HTML]{E4EEBC} & 63.78\cellcolor[HTML]{E4EEBC} \\
        \textbf{DepthForge(Ours)} & - & \textbf{66.19}\cellcolor[HTML]{BDE6CD} & \textbf{75.93}\cellcolor[HTML]{BDE6CD} & \textbf{67.24}\cellcolor[HTML]{BDE6CD} \\
        \bottomrule
    \end{tabular}
    }
    \caption{Performance comparison between our \textbf{DepthForge} and existing DGSS methods under the \textit{Citys. $\rightarrow$ BDD. + Map. + GTA5} generalization setting. Top three results are highlighted as \colorbox[HTML]{BDE6CD}{\textbf{best}}, \colorbox[HTML]{E4EEBC}{second}, and \colorbox[HTML]{FFF8C5}{third}, respectively. (\%)}
    \label{tab: C2BMG}
\end{table}

More importantly, the depth VFM itself possesses strong generalization capabilities, allowing domain adaptation without fine-tuning. A pre-trained VFM applies the same. Therefore, the overall optimation objective still focuses on the segmentation head $\mathcal{H}$ with parameter $\theta_{h}$ and our DepthForge with parameter $\theta_{D}$:
\begin{equation}
    \arg \underset{\theta_{D}, \theta_{h}} {min} \sum_{i=1}^{N} \mathcal{L}oss(\mathcal{H_{\theta_{h}}}(\mathcal{F}_{\Theta^{v}, \Theta^{d}, \theta_{D}}(x_{i})), y_{i}),
\end{equation}
where $x_{i}$ and $y_{i}$ denote the input image and its corresponding ground truth, respectively, and $N$ signifies the total number of images. $\mathcal{F}$ represent the forward process of our DepthForge strategy applied to the pre-trained VFMs with parameters $\Theta^{v}$ and $\Theta^{d}$.

\begin{figure*}[!t]
    \centering
    \setlength{\abovecaptionskip}{0.1cm}
    \setlength{\belowcaptionskip}{-0.3cm}
    \includegraphics[width=0.95\linewidth]{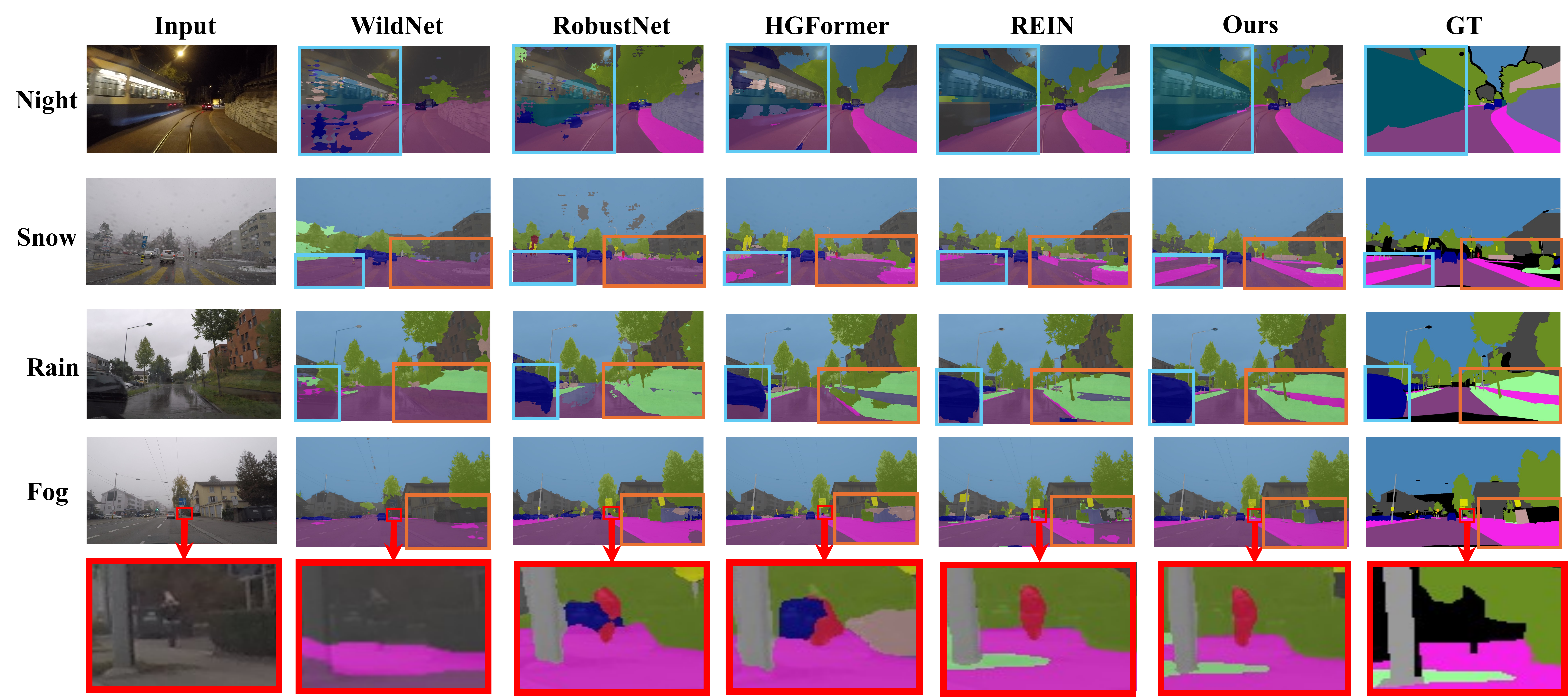}
    \caption{Key segmentation examples of existing DGSS methods and DepthForge under the Cityscapes to ACDC unseen target domains under Night, Snow, Rain, and Fog conditions. }
    \label{fig: C2A}
\end{figure*}

\begin{figure*}[!t]
    \centering
    \setlength{\abovecaptionskip}{0.1cm}
    \setlength{\belowcaptionskip}{-0.3cm}
    \includegraphics[width=0.95\linewidth]{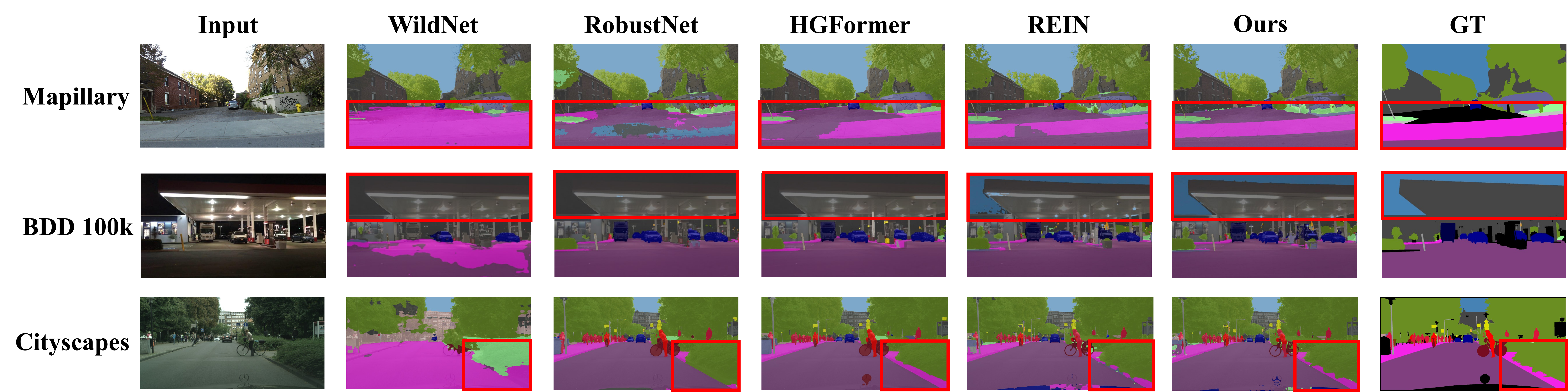}
    \caption{Key segmentation examples of existing DGSS methods and DepthForge under the GTA5 to Mapillary, BDD100k, and Cityscapes unseen target domains setting.}
    \label{fig: G2MBC}
\end{figure*}

\section{DepthForge}

This paper introduces the DepthForge strategy, a novel approach for domain generalized semantic segmentation. As shown in Fig.~\ref{fig: framework}, DepthForge consists of three main parts: (1) Depth Awareness; (2) Attention Optimization; (3) Depth Refinement Decoder. We detail the designs as follows.

\subsection{Depth Awareness}

Recent DGSS methods have applied pre-trained VFMs to source domain segmentation, with the core idea being to enhance the features between frozen VFM layers to help the model better adapt to target tasks. However, we observe that visual cues exhibit significant variations and even deficiencies across different scenes. Consequently, a more robust cue is required to provide a stronger prior for domain generalization. This motivates us to explore a strategy that focuses on the inherent similarities across different domains. In general, spatial location priors are able to help differentiate instances even when visual cues fail. To this end, we employ an additional depth VFM as an auxiliary network, as shown in Fig.~\ref{fig: framework}.

Specifically, given a VFM $\mathcal{V}$ and a depth VFM $\mathcal{D}$ with $N$ layers $\{L_{1}, L_{2}, ..., L_{N} \}$, the corresponding features are denoted as $\{f^{v}_{i}, f^{d}_{i} | f_{i} \in \mathbb{R}^{n \times c}, i \in [1, N] \}$. Note that the layers $\{L_{1}, L_{2}, ..., L_{N} \}$ are kept frozen. We acquire the visual feature $f_{i}^{v}$ from the $i$-th layer $L_{i}$ of $\mathcal{V}$, and the depth feature $f_{i}^{d}$ from the corresponding layer of $\mathcal{D}$. Our goal is to generate the enhanced feature $\Delta f_{i}=DepthForge(f^{v}_{i}, f^{d}_{i})$ by using both visual and depth cues.

We employ a set of learnable tokens $T = \{ T_{i} \in \mathbb{R}^{m \times c} | i \in [1, N] \}$ to adaptively learn scene knowledge. As we consider, relying solely on visual cues can be limiting and prone to failure. Therefore, we utilize visual and depth cues to construct learnable tokens, which we term depth-aware learnable tokens. We integrate $T_{i}$ with $f^{v}_{i}$ and $f^{d}_{i}$ using an attention-based mechanism, prompting tailored adjustments to bridge the gap between different domains.

We use $f^{v, d}_{i}$ as the $query$, and $T_{i}$ as the $key$ and the $value$ A dot-product operation is applied to generate awareness map $A^{v, d}_{i}$, and the $Softmax$ function ($S$) is employed to align each patch with a unique instance. The $A^{v, d}_{i}$ provides insights into both the visual cues and depth retrieval, qualitatively evaluating the relationship between various tokens and visual or depth cues, respectively. Subsequently, we incorporate depth awareness bias $A^{d}_{i}$ into the visual features $A^{v}_{i}$ to provide additional spatial relationships, thereby establishing a latent connection between pixels and the real world. The process is mathematically formulated as: 
\begin{equation}
    A_{i} = S(\frac{f^{v}_{i} \times (T^{v})^{\top}_{i}}{\sqrt{c}}) + \lambda  S(\frac{f^{d}_{i} \times (T^{d})^{\top}_{i}}{\sqrt{c}}).
\end{equation}

\subsection{Attention Optimization}

Directly leveraging the awareness representation $A_{i}$ to generate enhanced features may introduce a significant amount of irrelevant information for downstream tasks. To this end, we employ depth-aware learnable tokens $T_{i}$ to learn an alignment weight $W_{T_{i}}$ and bias $b_{T_{i}}$. The $T_{i}$ with weight $W_{T_{i}}$ and bias $b_{T_{i}}$ is then multiplied with the $A_{i}$ to obtain the initial enhanced features $\hat{\Delta f_{i}}$ that embed both visual and depth cues, using the equation:
\begin{equation}
    \hat{\Delta f_{i}} = A_{i} \times (T_{i} \times W_{T_{i}} + b_{T_{i}}),
\end{equation}
where we select the same optimization strategy as REIN, discarding high-weight features when the features are sufficiently precise to avoid unnecessary adjustments and mitigate the risk of inappropriate variations. Finally, we employ two consecutive residual connections $\phi$ to enhance the flexibility of attention feature adjustment and prevent gradient vanishing during the forward pass. The calculation is defined as $\Delta f_{i} = \phi \hat{\Delta f_{i}}$. Benefiting from these depth-aware and attention optimization adjustments, DepthForge is capable of generating more distinctive, discriminative features for different categories within a single image, while balancing both visual and depth cues.

\subsection{Depth Refinement Decoder}

The depth-aware features at different layers are inconsistent. Therefore, how to design a robust decoder becomes a core challenge that effectively integrate these enhanced features across various scales to accomplish downstream tasks.

Previous methods rely solely on the output of the final layer $f_{N}$ for semantic segmentation predictions. In contrast, we fuse the outputs from layers $L_{1} - L_{N}$ to refine the depth features. Specifically, we process these multi-layer features with an MLP with two different fully-connected layers $\phi$ and $\varphi$ and then concatenate them after a ReLU activation function:
\begin{equation}
    f^{out}_{i} = \phi (ReLU(\varphi f_{i})),
\end{equation}
\begin{equation}
    F_{u} = \gamma Concat(f^{out}_{1}, f^{out}_{2}, ..., f^{out}_{N}),
\end{equation}
where $\gamma$ is a convolutional layer. Finally, we design several sequential multi-head Transformer layers to generate the final predictions.

\begin{figure}
    \centering
    \setlength{\abovecaptionskip}{0.1cm}
    \setlength{\belowcaptionskip}{-0.3cm}
    \includegraphics[width=1\linewidth]{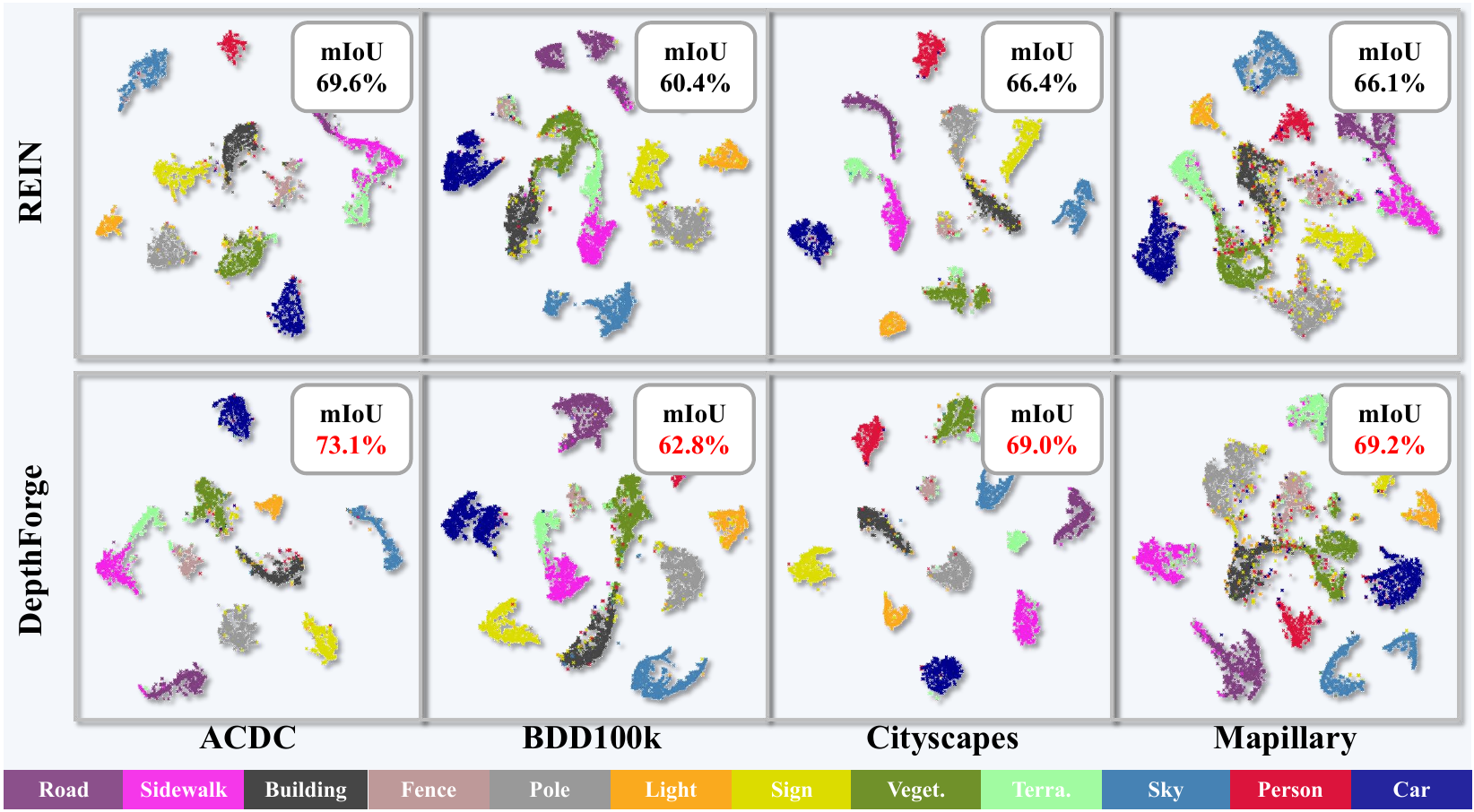}
    \caption{t-SNE visualization compares REIN and our \textbf{DepthForge}. The first volume corresponds to \textit{Citys. $\rightarrow$ ACDC}, and two through four volumes correspond to \textit{GTA5 $\rightarrow$ BDD. + Citys. + Map. } The best mIoU results are highlighted in \textbf{\textcolor{red}{red}}.}
    \label{fig: t-sne}
\end{figure}

\section{Experiment}

{
\setlength{\abovecaptionskip}{0.1cm}
\setlength{\belowcaptionskip}{-0.3cm}
\begin{table}[!t]
    \centering
    \scriptsize
    \setlength{\tabcolsep}{4pt}
    \renewcommand{\arraystretch}{1.1}
    \begin{tabular}{c|c|ccc}
        \toprule
        \textbf{Method} & \textbf{Proc. \& Year} & \textbf{Cityscapes} & \textbf{BDD100k} & \textbf{Mapillary} \\
        \midrule
        \multicolumn{2}{l}{\textit{ResNet based:}} \\
        DRPC~\cite{DRPC} & ICCV2019 & 37.42 & 32.14 & 34.12 \\
        GTR~\cite{gtrltr}  & TIP2021 & 37.53 & 33.75 & 34.52 \\
        RobustNet~\cite{ISW} & CVPR2021 & 36.58 & 35.20 & 40.33 \\
        DIRL~\cite{DIRL}  & AAAI2022 & 41.04 & 39.15 & 41.60 \\
        SHADE~\cite{SHADE} & ECCV2022 & 44.65 & 39.28 & 43.34 \\
        WildNet~\cite{WildNet}  & CVPR2023 & 44.62 & 38.42 & 46.09 \\
        SAW~\cite{SAW}  & CVPR2022 & 39.75 & 37.34 & 41.86 \\
        AdvStyle~\cite{AdvStyle}  & NeurIPS2022 & 39.62 & 35.54 & 37.00 \\
        SPC~\cite{SPC}  & CVPR2023 & 44.10 & 40.46 & 45.51 \\
        BlindNet~\cite{BlindNet}  & CVPR2024 & 45.72 & 41.32 & 47.08 \\
        \midrule
        \multicolumn{2}{l}{\textit{Transformer based:}} \\
        CMFormer~\cite{CMFormer} & AAAI2024 & 55.31 & 49.91 & 60.09 \\
        \midrule
        \multicolumn{2}{l}{\textit{VFM based:}} \\
        DIDEX~\cite{DIDEX} & WACV2024 & 62.00 & 54.30 & 63.00 \\
        REIN~\cite{REIN} & CVPR2024 & 66.40\cellcolor[HTML]{FFF8C5} & 60.40\cellcolor[HTML]{FFF8C5} & 66.10\cellcolor[HTML]{FFF8C5} \\
        FADA~\cite{FADA}  & NeurIPS2024 & 68.23\cellcolor[HTML]{E4EEBC} & 61.94\cellcolor[HTML]{E4EEBC} & 68.09\cellcolor[HTML]{E4EEBC} \\
        \textbf{DepthForge(Ours)}  & - & \textbf{69.04}\cellcolor[HTML]{BDE6CD} & \textbf{62.82}\cellcolor[HTML]{BDE6CD} & \textbf{69.22}\cellcolor[HTML]{BDE6CD} \\
        \bottomrule
    \end{tabular}
    \caption{Performance comparison between our \textbf{DepthForge} and existing DGSS methods under the \textit{GTA5 $\rightarrow$ Citys. + BDD. + Map.} generalization setting. Top three results are highlighted as \colorbox[HTML]{BDE6CD}{\textbf{best}}, \colorbox[HTML]{E4EEBC}{second}, and \colorbox[HTML]{FFF8C5}{third}, respectively. (\%)}
    \label{tab: G2CBM}
\end{table}
}

\subsection{Datasets \& Evaluation Protocols}

We evaluate our DepthForge on four real-world datasets (Cityscapes \cite{cityscapes}, BDD100K \cite{bdd100k}, Mapillary \cite{mapillary}, ACDC \cite{9711067}) and a synthetic dataset (GTA5 \cite{gtav}). \textbf{Cityscapes} (Citys.) is an autonomous driving dataset comprising 2,975 training images and 500 validation images, each with a resolution of $2048\times1024$. \textbf{BDD100K} (BDD.) and \textbf{Mapillary} (Map.) provide 1,000 and 2,000 validation images, with resolutions of $1280\times720$ and $1902\times1080$, respectively. \textbf{ACDC} offers 406 validation images captured under extreme conditions: namely at Night, Snow, Fog, and Rain, each with a resolution of $1902\times1080$. \textbf{GTA5} is a synthetic dataset that provides 24,966 annotated images obtained from the game.

Following the evaluation protocol of the existing DGSS methods, a certain dataset is chosen as the source domain for training, and the other three datasets serve as unseen target domains for validation. The three common evaluation settings are as follows: (1) Cityscapes $\rightarrow$ ACDC (Night, Snow, Fog, Rain); (2) GTA5 $\rightarrow$ Cityscapes, BDD100K, Mapillary; and (3) Cityscapes $\rightarrow$ BDD100K, Mapillary, GTA5. The evaluation metric is the mean Intersection-over-Union (mIoU).

\subsection{Deployment Details \& Parameter Settings}

Our implementation is built on the MMSegmentation \cite{mmseg2020} codebase. For optimization, we employ AdamW optimizer with an initial learning rate of 1e-4, a weight decay of 0.05, an epsilon of 1e-8, and beta values of (0.9, 0.999). A OneCycleLR scheduler is applied over 50,000 total steps with a maximum learning rate of 1e-4, featuring a warmup phase covering 10\% of the training iterations, and using cosine annealing, with both the div\_factor and final\_div\_factor set to 10. 

Data augmentation includes multi-scale resizing, random cropping (with a fixed crop size and category ratio constraints), random flipping, and photometric distortion. All experiments are conducted on an NVIDIA RTX 3090 GPU. 

\subsection{Comparison with State-of-The-Art Methods}

\begin{figure*}[!t]
    \centering
    \setlength{\abovecaptionskip}{0.1cm}
    \setlength{\belowcaptionskip}{-0.1cm}
    \includegraphics[width=1\linewidth]{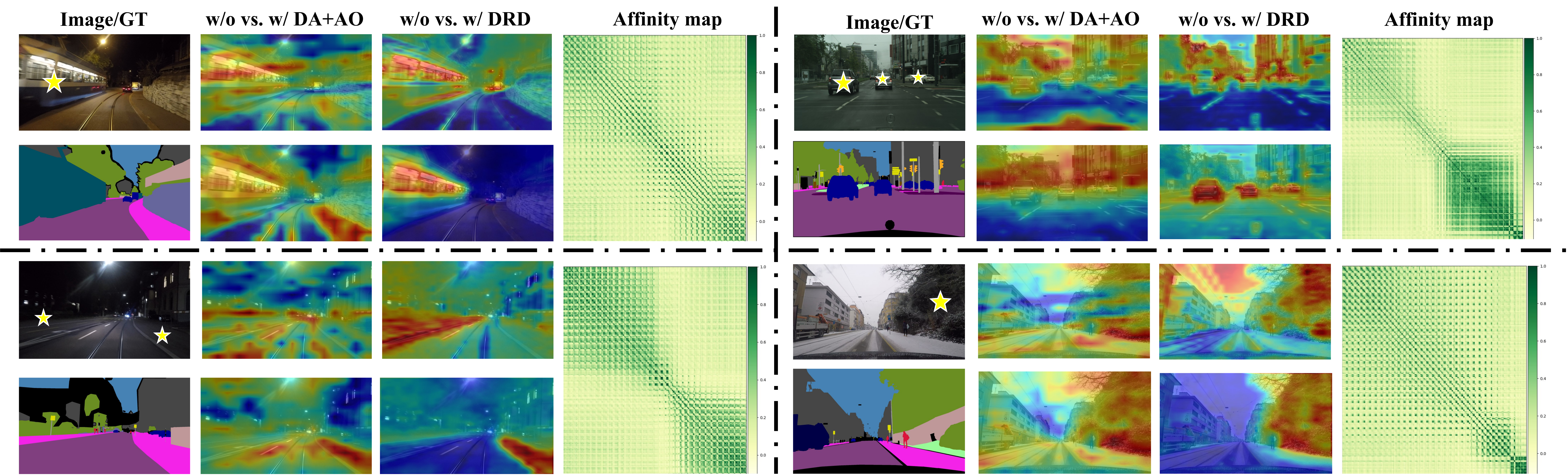}
    \caption{Visualizations of attention maps and affinity maps under differnet configurations, where DRD denotes the Depth Refinement Decoder, DA and AO represent the Depth Awareness and Attention Optimization, respectively.}
    \label{fig: affinity}
\end{figure*}

\begin{table}[!t]
    \centering
    \setlength{\abovecaptionskip}{0.1cm}
    \setlength{\belowcaptionskip}{-0.3cm}
    \resizebox{\linewidth}{!}{%
        \renewcommand\arraystretch{1.0}
        \setlength\tabcolsep{3.0pt}
        \begin{tabular}{l|l|cccccc}
            \hline
            Backbone & Configurations & \textbf{Snow} & \textbf{Night} & \textbf{Fog} & \textbf{Rain} & \textbf{BDD} & \textbf{Map} \\
            \hline
            \multirow{7}{*}{EVA02 (Large)~\cite{EVA,EVA02}} 
                & REIN~\cite{REIN}
                & 66.0& 52.1& 76.0& 69.5& 60.0& 70.0 \\
                & $+$ concat $f^{d}_{i}$
                & 66.3& 52.2& 76.2& 69.8& 60.2& 70.2 \\
                & $+$ Prompt Depth Anything
                & 65.8& 52.4& 75.5& 69.3& 60.4& 69.9 \\
                & $+$ Depth Anything V2
                & 67.2& 53.0& 77.0& 70.2& 60.9& 70.8 \\
                & $+$ DRD
                & 67.7\cellcolor[HTML]{FFF8C5}& 53.5\cellcolor[HTML]{FFF8C5}& 77.6\cellcolor[HTML]{FFF8C5}& 70.7\cellcolor[HTML]{FFF8C5}& 61.2\cellcolor[HTML]{FFF8C5}& 71.0\cellcolor[HTML]{FFF8C5} \\
                & $+$ DA + AO& 67.9\cellcolor[HTML]{E4EEBC}& 53.7\cellcolor[HTML]{E4EEBC}& 77.8\cellcolor[HTML]{E4EEBC}& 70.9\cellcolor[HTML]{E4EEBC}& 61.5\cellcolor[HTML]{E4EEBC}& 71.2\cellcolor[HTML]{E4EEBC} \\
                & \textbf{DepthForge}& \textbf{69.5}\cellcolor[HTML]{BDE6CD}& \textbf{54.5}\cellcolor[HTML]{BDE6CD}& \textbf{78.5}\cellcolor[HTML]{BDE6CD}& \textbf{72.0}\cellcolor[HTML]{BDE6CD}& \textbf{62.0}\cellcolor[HTML]{BDE6CD}& \textbf{73.0}\cellcolor[HTML]{BDE6CD} \\
            \hline
            \multirow{7}{*}{DINOv2 (Large)~\cite{DINOV2}} 
                & REIN~\cite{REIN}& 70.6& 55.9& 79.5& 72.5& 63.5& 74.0 \\
                & $+$ concat $f^{d}_{i}$& 70.8& 56.1& 79.4& 72.8& 63.6& 74.2 \\
                & $+$ Prompt Depth Anything& 70.4& 55.5& 79.7& 72.0& 63.8
                & 73.9 \\
                & $+$ Depth Anything V2& 71.5& 56.7& 80.5& 73.3& 64.4& 74.5 \\
                & $+$ DRD
                & 72.0\cellcolor[HTML]{FFF8C5}& 57.2\cellcolor[HTML]{FFF8C5}& 80.7\cellcolor[HTML]{FFF8C5}& 74.0\cellcolor[HTML]{FFF8C5}& 64.9\cellcolor[HTML]{FFF8C5}& 74.8\cellcolor[HTML]{FFF8C5} \\
                & $+$ DA + AO
                & 72.2\cellcolor[HTML]{E4EEBC}& 57.4\cellcolor[HTML]{E4EEBC}& 80.9\cellcolor[HTML]{E4EEBC}& 74.2\cellcolor[HTML]{E4EEBC}& 65.1\cellcolor[HTML]{E4EEBC}& 75.0\cellcolor[HTML]{E4EEBC} \\
                & \textbf{DepthForge}& \textbf{75.4}\cellcolor[HTML]{BDE6CD}& \textbf{60.4}\cellcolor[HTML]{BDE6CD}& \textbf{81.2}\cellcolor[HTML]{BDE6CD}& \textbf{75.4}\cellcolor[HTML]{BDE6CD}& \textbf{66.2}\cellcolor[HTML]{BDE6CD}& \textbf{76.0}\cellcolor[HTML]{BDE6CD} \\
            \hline
        \end{tabular}%
    }
    \caption{Ablation studies on component configurations of the proposed \textbf{DepthForge} under \textit{Citys. $\rightarrow$ ACDC with Snow, Night, Fog, Rain + BDD. + Map.} generalization setting, where DRD denotes the Depth Refinement Decoder, DA and AO represent the Depth Awareness and Attention Optimization. Top three results are highlighted as \colorbox[HTML]{BDE6CD}{\textbf{best}}, \colorbox[HTML]{E4EEBC}{second}, and \colorbox[HTML]{FFF8C5}{third}, respectively. (\%)} 
    \label{tab: ablation}
\end{table}

\begin{table}
\setlength{\abovecaptionskip}{0.1cm}
\setlength{\belowcaptionskip}{-0.3cm}
\centering
\resizebox{\linewidth}{!}{
\begin{tabular}{c|l|c|c|c|c|c}
\toprule
\textbf{No.} & \textbf{Design} & \textbf{Avg. Gain (\%)} & \textbf{Rain} & \textbf{Night} & \textbf{Snow} & \textbf{Rog} \\
\midrule
1 & No Depth & - & 70.6 & 55.9 & 79.5 & 72.5 \\
2 & Concatenation & -1.1 & 69.8 & 55.0 & 78.3 & 71.2 \\
\rowcolor{third}
3 & Depth Token & 1.6 & 72.3 & 57.8 & 80.7 & 74.0 \\
4 & Config1 & 0.2 & 70.8 & 56.1 & 79.4 & 72.8 \\
5 & Config2 & 0.7 & 71.5 & 56.8 & 80.1 & 73.0 \\
\rowcolor{best}
6 & Config3, \textbf{Ours} & \textbf{\textcolor{red}{3.5}} & \textbf{75.4} & \textbf{60.4} & \textbf{81.2} & \textbf{75.4} \\
\rowcolor{second}
7 & w/o Scale Factor & 2.8 & 74.8 & 59.3 & 80.9 & 74.8 \\
\bottomrule
\end{tabular}
}
\caption{Performance comparison of the different configurations. Top three results are highlighted as \colorbox[HTML]{BDE6CD}{\textbf{best}}, \colorbox[HTML]{E4EEBC}{second}, and \colorbox[HTML]{FFF8C5}{third}, respectively. (\%)}
\label{tab: config}
\end{table}

We compare our approach with existing DGSS methods, including: (1) ResNet-based methods, (2) Transformer-based methods, and (3) VFM-based methods.

\textbf{Cityscapes $\rightarrow$ ACDC}. Tab.~\ref{tab: C2A} presents the performance on the Cityscapes source domain and the ACDC target domain. Compared with the VFM-based REIN and FADA, the proposed DepthForge shows mIoU improvement of 4.8\% \& 1.9\%, 4.5\% \& 3.0\%, 1.7\% \& 1.0\%, and 2.9\% \& 0.4\% for the Snow, Night, Fog, and Rain scenes, respectively. The feature space of the REIN and the proposed method is visualized in the first volume of Fig.~\ref{fig: t-sne}. Moreover, our proposed method demonstrates superior performance compared to ResNet-based and Transformer-based approaches. These experimental results underscore the strong and robust generalization ability of DepthForge in extreme scenes with limited visual cues. Some exemplar segmentation results are compared under the Cityscapes $\rightarrow$ ACDC are provided in Fig.~\ref{fig: C2A}. Our DepthForge shows better pixel-wise predictions than previous methods, demonstrating the effectiveness and significance of depth awareness in our design. 

\begin{table}[tbp]
    \centering
    \setlength{\abovecaptionskip}{0.1cm}
    \setlength{\belowcaptionskip}{-0.3cm}
    \resizebox{\linewidth}{!}{%
        \renewcommand\arraystretch{1.1}
        \setlength\tabcolsep{3pt}
        \begin{tabular}{ll|c|ccc|c}
            \hline
            \rowcolor{white}
            \multirow{2}{*}{Backbone} & \multirow{2}{*}{\begin{tabular}[c]{@{}l@{}}Fine-tune\\ Method\end{tabular}} & \multirow{2}{*}{\begin{tabular}[c]{@{}c@{}}Trainable\\ Params$^*$\end{tabular}} & \multicolumn{3}{c|}{mIoU} & \multirow{2}{*}{Avg.} \\
            \cline{4-6}
                                      &        &          & Citys & BDD & Map &       \\
            \hline
            \multirow{3}{*}{\begin{tabular}[c]{@{}l@{}}CLIP~\cite{CLIP}\\ (ViT-Large)\end{tabular}}  
                                      & Full   & 304.15M  & 51.3  & 47.6  & 54.3  & 51.1\cellcolor[HTML]{EFEFEF}  \\
                                      & Freeze & ~~~0.00M & 53.7  & 48.7  & 55.0  & 52.4\cellcolor[HTML]{EFEFEF}  \\
                                      & REIN   & ~~~2.99M & 57.1\cellcolor[HTML]{FFF8C5}  & 54.7\cellcolor[HTML]{FFF8C5}  & 60.5\cellcolor[HTML]{FFF8C5}  & 57.4\cellcolor[HTML]{FFF8C5}  \\
                                      & FADA   & ~~~11.65M & 58.7\cellcolor[HTML]{E4EEBC}  & 55.8\cellcolor[HTML]{E4EEBC}  & 62.1\cellcolor[HTML]{E4EEBC}  & 58.9\cellcolor[HTML]{E4EEBC}  \\
                                      & \textbf{DepthForge(Ours) }  & ~~~2.99M & \textbf{59.7}\cellcolor[HTML]{BDE6CD} & \textbf{57.0}\cellcolor[HTML]{BDE6CD} & \textbf{63.5}\cellcolor[HTML]{BDE6CD}    & \textbf{60.0}\cellcolor[HTML]{BDE6CD}  \\
                                      
            \hline
            \multirow{3}{*}{\begin{tabular}[c]{@{}l@{}}SAM~\cite{SAM}\\ (Huge)\end{tabular}}  
                                      & Full   & 632.18M  & 57.6  & 51.7  & 61.5  & 56.9\cellcolor[HTML]{EFEFEF}  \\
                                      & Freeze & ~~~0.00M & 57.0  & 47.1  & 58.4  & 54.2\cellcolor[HTML]{EFEFEF}  \\
                                      & REIN   & ~~~4.51M & 59.6\cellcolor[HTML]{FFF8C5}  & 52.0\cellcolor[HTML]{FFF8C5}  & 62.1\cellcolor[HTML]{FFF8C5}  & 57.9\cellcolor[HTML]{FFF8C5}  \\
                                      & FADA   & ~~~16.59M & 61.0\cellcolor[HTML]{E4EEBC}  & 53.2\cellcolor[HTML]{E4EEBC}  & 63.4\cellcolor[HTML]{E4EEBC}  & 60.0\cellcolor[HTML]{E4EEBC}  \\
                                      & \textbf{DepthForge(Ours)}   & ~~~4.51M & \textbf{63.3}\cellcolor[HTML]{BDE6CD} & \textbf{55.5}\cellcolor[HTML]{BDE6CD} & \textbf{65.1}\cellcolor[HTML]{BDE6CD}    & \textbf{61.3}\cellcolor[HTML]{BDE6CD}  \\
            \hline
            \multirow{3}{*}{\begin{tabular}[c]{@{}l@{}}EVA02~\cite{EVA,EVA02}\\ (Large)\end{tabular}}  
                                      & Full   & 304.24M  & 62.1  & 56.2  & 64.6  & 60.9\cellcolor[HTML]{EFEFEF}  \\
                                      & Freeze & ~~~0.00M & 56.5  & 53.6  & 58.6  & 56.2\cellcolor[HTML]{EFEFEF}  \\
                                      & REIN   & ~~~2.99M & 65.3\cellcolor[HTML]{FFF8C5}  & 60.5\cellcolor[HTML]{FFF8C5}  & 64.9\cellcolor[HTML]{FFF8C5}  & 63.6\cellcolor[HTML]{FFF8C5}  \\
                                      & FADA   & ~~~11.65M & 66.7\cellcolor[HTML]{E4EEBC}  & \textbf{61.9}\cellcolor[HTML]{BDE6CD}  & 66.1\cellcolor[HTML]{E4EEBC}  & 64.9\cellcolor[HTML]{E4EEBC}  \\
                                      & \textbf{DepthForge(Ours)}  & ~~~2.99M & \textbf{68.0}\cellcolor[HTML]{BDE6CD} & 61.7\cellcolor[HTML]{E4EEBC} & \textbf{67.5}\cellcolor[HTML]{BDE6CD}    & \textbf{65.7}\cellcolor[HTML]{BDE6CD}   \\
            \hline
            \multirow{3}{*}{\begin{tabular}[c]{@{}l@{}}DINOV2~\cite{DINOV2}\\ (Large)\end{tabular}}  
                                      & Full   & 304.20M  & 63.7  & 57.4  & 64.2  & 61.7\cellcolor[HTML]{EFEFEF}  \\
                                      & Freeze & ~~~0.00M & 63.3  & 56.1  & 63.9  & 61.1\cellcolor[HTML]{EFEFEF}  \\
                                      & REIN   & ~~~2.99M & 66.4\cellcolor[HTML]{FFF8C5}  & 60.4\cellcolor[HTML]{FFF8C5}  & 66.1\cellcolor[HTML]{FFF8C5}  & 64.3\cellcolor[HTML]{FFF8C5}  \\
                                      & FADA   & ~~~11.65M & 68.2\cellcolor[HTML]{E4EEBC}  & 62.0\cellcolor[HTML]{E4EEBC} & 68.1\cellcolor[HTML]{E4EEBC}  & 66.1\cellcolor[HTML]{E4EEBC}  \\
                                      & \textbf{DepthForge(Ours)}   & ~~~2.99M & \textbf{69.0}\cellcolor[HTML]{BDE6CD} & \textbf{62.8}\cellcolor[HTML]{BDE6CD} & \textbf{69.2}\cellcolor[HTML]{BDE6CD}    & \textbf{67.0}\cellcolor[HTML]{BDE6CD}  \\
            \hline
        \end{tabular}
    }
    \caption{Performance comparison of the proposed \textbf{DepthForge} across multiple VFMs as backbones under the \textit{GTA5 $\rightarrow$ Citys. + BDD. + Map.} generalization setting. Mark $^*$ denotes trainable parameters in backbones. Top three results are highlighted as \colorbox[HTML]{BDE6CD}{\textbf{best}}, \colorbox[HTML]{E4EEBC}{second}, and \colorbox[HTML]{FFF8C5}{third}, respectively. (\%)}
    \label{tab: VFMs}
\end{table}

\begin{table}[tbp]
    \centering
    \setlength{\abovecaptionskip}{0.1cm}
    \setlength{\belowcaptionskip}{-0.3cm}
    \resizebox{\linewidth}{!}{%
        \renewcommand\arraystretch{1.1}
        \setlength\tabcolsep{3pt}
        \begin{tabular}{ll|c|ccc|c}
            \hline
            \rowcolor{white}
            \multirow{2}{*}{Backbone} & \multirow{2}{*}{Fine-tune Method} & \multirow{2}{*}{\begin{tabular}[c]{@{}c@{}}Trainable\\ Params$^*$\end{tabular}} & \multicolumn{3}{c|}{mIoU} & \multirow{2}{*}{Avg.} \\
            \cline{4-6}
                                      &                                   &                                     & Citys & BDD & Map &       \\
            \hline
            \multirow{10}{*}{\begin{tabular}[c]{@{}l@{}}EVA02\\ (Large)~\cite{EVA,EVA02}\end{tabular}} 
                                      & Full                              & 304.24M                            & 62.1  & 56.2  & 64.6  & 60.9\cellcolor[HTML]{EFEFEF} \\
                                      & +AdvStyle~\cite{AdvStyle}         & 304.24M                            & 63.1  & 56.4  & 64.0  & 61.2\cellcolor[HTML]{EFEFEF} \\
                                      & +PASTA~\cite{PASTA}               & 304.24M                            & 61.8  & 57.1  & 63.6  & 60.8\cellcolor[HTML]{EFEFEF} \\
                                      & +GTR-LTR~\cite{gtrltr}            & 304.24M                            & 59.8  & 57.4  & 63.2  & 60.1\cellcolor[HTML]{EFEFEF} \\
            \cline{2-7}
                                      & Freeze                            & 0.00M                              & 56.5  & 53.6  & 58.6  & 56.2\cellcolor[HTML]{EFEFEF} \\
                                      & +AdvStyle~\cite{AdvStyle}         & 0.00M                              & 51.4  & 51.6  & 56.5  & 53.2\cellcolor[HTML]{EFEFEF} \\
                                      & +PASTA~\cite{PASTA}               & 0.00M                              & 57.8  & 52.3  & 58.5  & 56.2\cellcolor[HTML]{EFEFEF} \\
                                      & +GTR-LTR~\cite{gtrltr}            & 0.00M                              & 52.5  & 52.8  & 57.1  & 54.1\cellcolor[HTML]{EFEFEF} \\
                                      & +LoRA~\cite{lora}                 & 1.18M                              & 55.5  & 52.7  & 58.3  & 55.5\cellcolor[HTML]{EFEFEF} \\
                                      & +AdaptFormer~\cite{adaptformer}   & 3.17M                              & 63.7  & 59.9  & 64.2  & 62.6\cellcolor[HTML]{EFEFEF} \\
                                      & +VPT~\cite{vpt}                   & 3.69M                              & 62.2  & 57.7  & 62.5  & 60.8\cellcolor[HTML]{EFEFEF} \\
                                      & +REIN~\cite{REIN}                 & 2.99M                              & 65.3\cellcolor[HTML]{FFF8C5} & 60.5\cellcolor[HTML]{FFF8C5} & 64.9\cellcolor[HTML]{FFF8C5} & 63.6\cellcolor[HTML]{FFF8C5} \\
                                      & +FADA~\cite{FADA}                 & 11.65M                             & 66.7\cellcolor[HTML]{E4EEBC} & \textbf{61.9}\cellcolor[HTML]{BDE6CD} & 66.1\cellcolor[HTML]{E4EEBC} & 64.9\cellcolor[HTML]{E4EEBC} \\
                                      & \textbf{+DepthForge(Ours)}               & 2.99M                              & \textbf{68.0}\cellcolor[HTML]{BDE6CD} & 61.7\cellcolor[HTML]{E4EEBC} & \textbf{67.5}\cellcolor[HTML]{BDE6CD}    & \textbf{65.7}\cellcolor[HTML]{BDE6CD} \\
            \hline
            \multirow{10}{*}{\begin{tabular}[c]{@{}l@{}}DINOv2\\ (Large)~\cite{DINOV2}\end{tabular}} 
                                      & Full                              & 304.20M                            & 63.7  & 57.4  & 64.2  & 61.7\cellcolor[HTML]{EFEFEF} \\
                                      & +AdvStyle~\cite{AdvStyle}         & 304.20M                            & 60.8  & 58.0  & 62.5  & 60.4\cellcolor[HTML]{EFEFEF} \\
                                      & +PASTA~\cite{PASTA}               & 304.20M                            & 62.5  & 57.2  & 64.7  & 61.5\cellcolor[HTML]{EFEFEF} \\
                                      & +GTR-LTR~\cite{gtrltr}             & 304.20M                            & 62.7  & 57.4  & 64.5  & 61.6\cellcolor[HTML]{EFEFEF} \\
            \cline{2-7}
                                      & Freeze                            & 0.00M                              & 63.3  & 56.1  & 63.9  & 61.1\cellcolor[HTML]{EFEFEF} \\
                                      & +AdvStyle~\cite{AdvStyle}         & 0.00M                              & 61.5  & 55.1  & 63.9  & 60.1\cellcolor[HTML]{EFEFEF} \\
                                      & +PASTA~\cite{PASTA}               & 0.00M                              & 62.1  & 57.2  & 64.5  & 61.3\cellcolor[HTML]{EFEFEF} \\
                                      & +GTR-LTR~\cite{gtrltr}             & 0.00M                              & 60.2  & 57.7  & 62.2  & 60.0\cellcolor[HTML]{EFEFEF} \\
                                      & +LoRA~\cite{lora}                 & 0.79M                              & 65.2  & 58.3  & 64.6  & 62.7\cellcolor[HTML]{EFEFEF} \\
                                      & +AdaptFormer~\cite{adaptformer}   & 3.17M                              & 64.9  & 59.0  & 64.2  & 62.7\cellcolor[HTML]{EFEFEF} \\
                                      & +VPT~\cite{vpt}                   & 3.69M                              & 65.2  & 59.4  & 65.5  & 63.3\cellcolor[HTML]{EFEFEF} \\
                                      & +REIN~\cite{REIN}                 & 2.99M                              & 66.4\cellcolor[HTML]{FFF8C5} & 60.4\cellcolor[HTML]{FFF8C5} & 66.1\cellcolor[HTML]{FFF8C5} & 64.3\cellcolor[HTML]{FFF8C5} \\
                                      & +FADA~\cite{FADA}                 & 11.65M                             & 68.2\cellcolor[HTML]{E4EEBC} & 62.0\cellcolor[HTML]{E4EEBC} & 68.1\cellcolor[HTML]{E4EEBC} & 66.1\cellcolor[HTML]{E4EEBC} \\
                                      & \textbf{+DepthForge(Ours)}                 & 2.99M                              & \textbf{69.0}\cellcolor[HTML]{BDE6CD} & \textbf{62.8}\cellcolor[HTML]{BDE6CD} & \textbf{69.2}\cellcolor[HTML]{BDE6CD}    & \textbf{67.0}\cellcolor[HTML]{BDE6CD} \\
            \hline
        \end{tabular}
    }
    \caption{Performance comparison of the proposed \textbf{DepthForge} against other DGSS methods under the \textit{GTA5 $\rightarrow$ Citys. + BDD. + Map.} generalization setting. Mark $^*$ denotes trainable parameters in backbones. Top three results are highlighted as \colorbox[HTML]{BDE6CD}{\textbf{best}}, \colorbox[HTML]{E4EEBC}{second}, and \colorbox[HTML]{FFF8C5}{third}, respectively. (\%)}
    \label{tab: VFM+}
\end{table}

\textbf{Cityscapes $\rightarrow$ BDD, Map, GTA5}. Tab.~\ref{tab: C2BMG} shows that the proposed DepthForge achieves the state-of-the-art performance, outperforming the REIN and FADA by 2.65\% \& 1.07\%, 1.9\% \& 0.07\%, and 3.12\% \& 1.13\% on the BDD100k, Mapillary, and GTA unseen target domains, respectively. The proposed method achieves an improvement of approximately 10\% compared to ResNet- and Transformer-based approaches.

\textbf{GTA5 $\rightarrow$ Citys, Map, BDD}. As shown in Tab.~\ref{tab: G2CBM}, the proposed DepthForge achieves mIoU improvements of 2.64\%, 2.42\%, and 3.12\% on the Cityscapes, BDD100k, and Mapillary unseen target domains, respectively. The feature space of REIN and our DepthForge is visualized in Fig.~\ref{fig: t-sne}, demonstrating the effectiveness of our depth-aware learnable tokens in the visual-spatial attention learning. In addition, we outperform the latest FADA by approximately 1\% across all three target domains. Visual examples for qualitative comparison are given in Fig.~\ref{fig: G2MBC}.

\subsection{Ablation Studies}

\textbf{Depth awareness and attention optimization are significant for depth-aware learnable tokens.} We find that simply incorporating a spatial cue bias into the original learnable tokens ($+f^{d}_{i}$) makes almost no difference. However, when we transform the original learnable tokens into depth-aware learnable tokens using the DepthForge framework, we achieve an approximate 2\% improvement in the target domains ACDC and Mapillary, as shown in Tab.~\ref{tab: ablation}. The depth-aware learnable tokens compensate for the shortcomings of visual cues and establish more robust visual-spatial attention. Compared to different configurations, our design allows gradients to be backpropagated through the tokens alone, enables each modality to contribute independently based on its own feature representations, preventing dominance from one during optimization, as shown in Tab.~\ref{tab: config} and Fig.~\ref{fig: loss}. This conclusion is supported by qualitative visualizations in Fig.~\ref{fig: affinity}.

\textbf{Depth refinement decoder facilitates the perception of both visual and depth information.} As shown in Tab.~\ref{tab: ablation}, we observe that depth refinement decoder integrates multi-scale information to enhance the expressive capacity of both the frozen VFM features and the enhanced features. With the incorporation of depth information, this optimization effect is further amplified. This conclusion is corroborated by qualitative visualizations in Fig.~\ref{fig: affinity}.

\subsection{Generalization on Other Setting}

\textbf{To Different Depth VFMs}. We evaluate the performance of the proposed DepthForge across different depth VFMs. Tab.~\ref{tab: ablation} shows the relative spatial relationships derived from Depth Anything v2 outperform the absolute spatial cues provided by Prompt Depth Anything. We believe that what is more crucial in an image is the relative spatial relationship, which establishes connections among pixels and between pixels and the world. In contrast, absolute spatial information introduces significant discrepancies that can act as noise to visual cues.

\textbf{To Different VFMs}. We evaluate the generalization ability of the proposed DepthForge across different VFMs. Tab.~\ref{tab: VFMs} shows the superiority of DepthForge when embedded into these VFMs.

\textbf{To Token Fine-Tuning Methods}. We further compare the proposed method with some existing fine-tuning methods. As shown in Tab.~\ref{tab: VFM+}, the proposed DepthForge outperforms these methods in all unseen target domains.

\section{Conclusion}

This paper presents the DepthForge framework, a novel approach for robust domain generalization semantic segmentation. We provide a detailed analysis of the challenges and limitations faced by existing methods. To this end, we propose DepthForge to integrate a pretrained depth VFM for fine-tuning. We design depth-aware learnable tokens to provide supplementary spatial cues to the visual features, thereby improving geometric consistency and enhancing the pixel discriminability. Moreover, we incorporate multi-layer enhanced features to effectively capture spatial correlations across various scales. Extensive experiments demonstrate the substantial potential of leveraging multi-modal VFMs in the field of DGSS, particularly under extreme conditions where visual cues are limited or missing, validating the effectiveness of our DepthForge in forging VFMs for DGSS.


\section*{Acknowledgements}

This work was supported in part by the Natural Science Foundation of Xiamen, China, under Grant 3502Z202373036; in part by the National Natural Science Foundation of China under Grant 42371457; in part by the Key Project of Natural Science Foundation of Fujian Province, China, under Grant 2022J02045.


{
    \small
    \bibliographystyle{ieeenat_fullname}
    \bibliography{main}
}

\end{document}